\def\year{2019}\relax
\documentclass[letterpaper]{article} 
\usepackage{aaai19}  
\usepackage{times}  
\usepackage{helvet}  
\usepackage{courier}  
\usepackage{url}  
\usepackage{graphicx}  
\usepackage{subfigure}

\usepackage{soul}
\usepackage[utf8]{inputenc}
\usepackage[small]{caption}
\usepackage{url} 
\usepackage{latexsym,amssymb,amsmath,amsthm}
\usepackage{xspace}

\usepackage{color}

\begin{document}

\newcommand{\myi}{(\emph{i})\xspace}
\newcommand{\myii}{(\emph{ii})\xspace}
\newcommand{\myiii}{(\emph{iii})\xspace}
\newcommand{\myiv}{(\emph{iv})\xspace}
\newcommand{\myv}{(\emph{v})\xspace}
\newcommand{\myvi}{(\emph{vi})\xspace}
\newcommand{\myvii}{(\emph{vii})\xspace}
\newcommand{\myviii}{(\emph{viii})\xspace}

\newcommand{\A}{\mathcal{A}} \newcommand{\B}{\mathcal{B}}
\newcommand{\C}{\mathcal{C}} \newcommand{\D}{\mathcal{D}}
\newcommand{\E}{\mathcal{E}} \newcommand{\F}{\mathcal{F}}
\newcommand{\G}{\mathcal{G}} \renewcommand{\H}{\mathcal{H}}
\newcommand{\I}{\mathcal{I}} \newcommand{\J}{\mathcal{J}}
\newcommand{\K}{\mathcal{K}} \renewcommand{\L}{\mathcal{L}}
\newcommand{\M}{\mathcal{M}} \newcommand{\N}{\mathcal{N}}
\renewcommand{\O}{\mathcal{O}} \renewcommand{\P}{\mathcal{P}}
\newcommand{\Q}{\mathcal{Q}} \newcommand{\R}{\mathcal{R}}
\renewcommand{\S}{\mathcal{S}} \newcommand{\T}{\mathcal{T}}
\newcommand{\U}{\mathcal{U}} \newcommand{\V}{\mathcal{V}}
\newcommand{\W}{\mathcal{W}} \newcommand{\X}{\mathcal{X}}
\newcommand{\Y}{\mathcal{Y}} \newcommand{\Z}{\mathcal{Z}}

\newcommand{\limp}{\mathbin{\rightarrow}}
\newcommand{\incl}{\subseteq}
\newcommand{\ind}{\hspace*{.18in}}

\newcommand{\Next}{\raisebox{-0.27ex}{\LARGE$\circ$}}
\newcommand{\Wnext}{\raisebox{-0.27ex}{\LARGE$\bullet$}}
\newcommand{\Until}{\mathop{\U}}
\newcommand{\Since}{\mathop{\S}}
\newcommand{\Release}{\mathop{\R}}
\newcommand{\Wuntil}{\mathop{\W}}
\newcommand{\true}{\mathit{true}}
\newcommand{\final}{\mathit{Final}}
\newcommand{\false}{\mathit{false}}
\newcommand{\ttrue}{{\mathit{tt}}}
\newcommand{\ffalse}{\mathit{ff}}
\newcommand{\Last}{\mathit{last}}
\newcommand{\Ended}{\mathit{end}}
\newcommand{\length}{\mathit{length}}
\newcommand{\last}{\mathit{n}}
\newcommand{\nnf}{\mathit{nnf}}
\newcommand{\CL}{\mathit{CL}}
\newcommand{\MU}[2]{\mu #1.#2}
\newcommand{\NU}[2]{\nu #1.#2}
\newcommand{\BOX}[1]{ [#1]}
\newcommand{\DIAM}[1]{\langle #1 \rangle}
\newcommand{\transl}{f}
\newcommand{\Int}[2][\I]{#2^{#1}}
\newcommand{\INT}[2][\I]{(#2)^{#1}}
\newcommand{\Inta}[2][\rho]{#2_{#1}^\I}
\newcommand{\INTA}[2][\rho]{(#2)_{#1}^\I}

\newcommand{\prev}{{\circleddash}}
\newcommand{\gpast}{{\boxminus}}
\newcommand{\past}{{\diamondminus}}
\newcommand{\since}{\mathop{\S}}

\newcommand{\LT}{{\sc lt}$_f$\xspace}
\newcommand{\LTi}{{\sc lt}$_i$\xspace}
\newcommand{\LTL}{{\sc ltl}\xspace}
\newcommand{\LTLf}{{\sc ltl}$_f$\xspace}
\newcommand{\PLTL}{{\sc pltl}\xspace}
\newcommand{\DFLTL}{$\$${\sc fltl}\xspace}
\newcommand{\LTLi}{{\sc ltl}\xspace}
\newcommand{\LDL}{{\sc ldl}\xspace}
\newcommand{\LDLf}{{\sc ldl}$_f$\xspace}
\newcommand{\RE}{{\sc re}$_f$\xspace}
\newcommand{\REGEX}{{\sc re}\xspace}
\newcommand{\PDL}{{\sc pdl}\xspace}
\newcommand{\FOf}{{\sc fo}$_f$\xspace}
\newcommand{\MSOf}{{\sc mso}$_f$\xspace}
\newcommand{\FO}{{\sc fo}\xspace}
\newcommand{\MSO}{{\sc mso}\xspace}
\newcommand{\AFW}{{\sc afw}\xspace}
\newcommand{\NFA}{{\sc nfa}\xspace}
\newcommand{\DFA}{{\sc dfa}\xspace}
\newcommand{\DFAs}{{\sc dfa}s\xspace}
\newcommand{\CTL}{{\sc ctl}\xspace}
\newcommand{\QLTL}{{\sc qltl}\xspace}
\newcommand{\muLTL}{$\mu${\sc ltl}\xspace}
\newcommand{\declare}{{\sc declare}\xspace}
\newcommand{\fol}{\mathit{fol}}
\newcommand{\f}{\mathit{f}}
\newcommand{\g}{\mathit{g}}
\newcommand{\re}{\mathit{re}}

\newcommand{\MDP}{{\sc mdp}\xspace}

\newcommand{\satf}{\ensuremath{\textsc{sat}_f}}
\newcommand{\sati}{\ensuremath{\textsc{sat}_i}}

\newcommand{\buchi}{B\"uchi\xspace}
\newcommand{\Nat}{{\rm I\kern-.23em N}}
\newcommand{\Prop}{\P}
\newcommand{\Var}{\V}

\newcommand{\tup}[1]{\langle #1 \rangle}

\newcommand{\Stop}{\mathit{stop}}
\newcommand{\rew}{\mathit{rew}}
\newcommand{\Tr}{\mathit{Tr}}

\newcommand{\pref}{\mathsf{pref}}
\newcommand{\temptrue}{\mathit{temp\_true}}
\newcommand{\willtemptrue}{\mathit{will\_temp\_true}}
\newcommand{\willtrue}{\mathit{will\_true}}
\newcommand{\willfalse}{\mathit{will\_false}}
\newcommand{\tempfalse}{\mathit{temp\_false}}
\newcommand{\possgood}{\mathit{poss\_good}}
\newcommand{\possbad}{\mathit{poss\_bad}}
\newcommand{\necgood}{\mathit{nec\_good}}
\newcommand{\necbad}{\mathit{nec\_bad}}

\newcommand{\LOGSPACE}{{\sc logspace}\xspace}
\newcommand{\NLOGSPACE}{{\sc nlogspace}\xspace}
\newcommand{\PTIME}{{\sc ptime}\xspace}
\newcommand{\NP}{{\sc np}\xspace}
\newcommand{\EXPTIME}{{\sc exptime}\xspace}
\newcommand{\PSPACE}{{\sc pspace}\xspace}
\newcommand{\TWOEXPTIME}{{\sc 2exptime}\xspace}

\newcommand{\expand}{\textbf{\textit{E}}}
\newcommand{\ttt}{{\textbf{\textit{T}}}}
\newcommand{\fff}{{\textbf{\textit{\texttt{F}}}}}

\newcommand{\fstate}{s_f}

\newcommand{\atomize}[1]{\texttt{"}\ensuremath{#1}\texttt{"}}

\newcommand{\Sapientino}{{\sc Sapientino}\xspace}
\newcommand{\Breakout}{{\sc Breakout}\xspace}
\newcommand{\Minecraft}{{\sc Minecraft}\xspace}
\newcommand{\CocktailParty}{{\sc CocktailParty}\xspace}
\newcommand{\Omni}{{\sc omni}\xspace}
\newcommand{\Differential}{{\sc differential}\xspace}
\newcommand{\Pose}{{\sc poseonly}\xspace}
\newcommand{\PoseColor}{{\sc posecolor}\xspace}

\newcommand{\OP}{{\sc OP}\xspace}
\newcommand{\OPC}{{\sc OPC}\xspace}
\newcommand{\DP}{{\sc DP}\xspace}
\newcommand{\DPC}{{\sc DPC}\xspace}

\newcommand{\Move}{{\sc move}\xspace}
\newcommand{\Fire}{{\sc fire}\xspace}
\newcommand{\Local}{{\sc local}\xspace}
\newcommand{\Global}{{\sc global}\xspace}

\newtheorem{theorem}{Theorem}
\newtheorem{lemma}[theorem]{Lemma}
\newtheorem{definition}{Definition}


%
\title{Foundations for Restraining Bolts:\\ Reinforcement Learning with LTLf/LDLf restraining specifications}
\author{Giuseppe De Giacomo \and
Luca Iocchi \and Marco Favorito \and   Fabio Patrizi\\
DIAG - Universit\`a di Roma ``La Sapienza'', Italy\\
\emph{\{lastname\}}@diag.uniroma1.it}
\maketitle

\begin{abstract}
  In this work we investigate on the concept of ``\emph{restraining bolt}'', envisioned in Science Fiction. Specifically we introduce a novel problem in AI. We have two distinct sets of features extracted from the world, one by the agent  and one by the authority imposing restraining specifications (the ``restraining bolt''). The two sets are apparently unrelated since of interest to independent parties, however they both account for (aspects of) the same world. We consider the case in which the agent is a reinforcement learning agent on the first set of features, while the restraining bolt is specified logically using linear time logic on finite traces \LTLf/\LDLf over the second set of features. 
We show formally, and illustrate with examples, that, under general circumstances, the agent can learn while shaping its goals to suitably conform  (as much as possible) to the restraining bolt specifications.
\end{abstract}

 \section{Introduction}

This work starts a scientific investigation on the
concept of ``\emph{restraining bolt}'', as envisioned in Science
Fiction. A restraining bolt is a
  ``device that restricts a droid's [agent's] actions when connected
  to its systems. Droid owners install restraining bolts to limit
  actions to a set of desired behaviors.''\footnote{
  \url{https://www.starwars.com/databank/restraining-bolt}} %
The concept of restraining bolt introduces a new problem in AI.  We
have two distinct representations of the world, one by the agent and
one by the authority imposing restraining specifications, i.e., the
bolt.  Such representations are apparently unrelated as developed by
independent parties, but both model (aspects of) the same
world.  We want the agent to conform (as much as possible) to the restraining specifications,
even if these are not expressed in terms of the agent's world representation.

Studying this problem from a classical Knowledge Representation
perspective \cite{Reiter01} would require to establish some sort of
``glue'' between the representation by the agent and that by the
restraining bolt.  Instead, we bypass dealing with such a ``glue'' by
studying this problem in the context of Reinforcement Learning (RL)
\cite{Puterman94,SuttonBarto98}, which is currently of great interest
to develop components with forms of decision making, possibly coupled
with deep learning techniques \cite{MnihKSRVBGRFOPB15,AlphaGoZero17}.

Specifically, we consider an agent and a restraining bolt of 
different nature.  The \emph{agent} is a reinforcement learning agent whose
``model'' of the world is a hidden, factorized, Markov Decision
Processes (MDP) over a certain set of world features. That is, the
state is factorized in a set of features observable to the agent,
while transition function and reward function are hidden.
The \emph{restraining bolt} consists in a logical specification of
traces that are considered desirable. The world features that are used
represent states in these traces are disjoint from those used by the
agent. More concretely such specifications are expressed in
full-fledged temporal logics over finite traces, \LTLf and its
extension \LDLf \cite{DegVa13,DeGiacomoS18,BDP-AAAI18}. Notice that
the restraining bolt does not have an explicit model of the dynamics
of the world, nor of the agent. Still it can assess if a given trace
generated by the execution of the agent in the world is desiderabile,
and give additional rewards when it does.

The connection between the agent and the restraining bolt is 
loose: the bolt provides additional reward to the agent
and only needs to know the order of magnitude of the original rewards of
the agent to suitably fix a scaling factor\footnote{Note that finding the right scaling factor is an importan issue in RL \cite{SimsekB06}, but out of the scope of the paper.} for its own additional
rewards. In addition, it provides to the agent additional
features to allow the agent to know at what stage of the satisfaction
of temporal formulas the world is so that the agent can choose its
policy accordingly. Without them, the agent would not be able to act
differently at different stages to get the rewards according to the
temporal specifications.

The main result of this paper is that, in spite of the loose connection
between the two models, under general circumstances, the \emph{agent can
learn to act so as to conform as much as possible to the \LTLf/\LDLf
specifications}.
Observe that we deal with two separate representations (i.e.,
two distinct sets of features), one for the agent and one for the
bolt, which are apparently unrelated, but in reality, correlated by
the world itself, cf., \cite{Brooks91}.  The crucial point is that, in
order to perform RL effectively in presence of a restraining bolt
\emph{such a correlation does not need to be formalized}.




For example, consider a service robot serving drinks and snacks at a party. The robot knows the locations where it can grasp drink and snack items and the locations of people that can receive such items. The robot can execute actions to move in the environment, to grasp objects and to deliver them to people. With rewards associated to effective deliver, the robot can learn how to serve something to a specific person.
Now, suppose we want to impose the following restraining bolt specification:
\emph{serve exactly one drink and one snack to every person, and do
  not serve alcoholic drinks to minors}. To express this specification (e.g., as
an \LTLf/\LDLf formula), a representation of the status of each person
(i.e., identity, age and received items) is needed, even though these
features are \emph{not} available to the learning agent (but only to
the restraining bolt).
%
%

Notice that the presence of \LTLf/\LDLf specification
makes the whole system formed by the agent and the restraining
bolt non-Markovian.
Recently, interest in non-Markovian Reward Decision Processes
NMRDPs
\cite{BBG96,WHITEHEAD1995271}, and in particular on expressing rewards using linear-time temporal logic has been
revived and motivated by the difficulty in rewarding complex behaviors
directly on MDPs \cite{Littman15,LittmanTFIWM17}.
In this context, the use of linear time logics over finite traces such as \LTLf or its extension \LDLf  has been recently advocated 
\cite{CamachoCSM17b,BDP-AAAI18,IcarteKVM18}.
Notably, both \LTLf and \LDLf formulas can
be transformed into deterministic finite state automata tracking the
stage of satisfaction of the formulas \cite{DegVa13}. This, in turn,
allows for transforming an NMRDP with non-Markovian \LTLf/\LDLf rewards
into an equivalent MDP over an extended state space, obtained as the
crossproduct of the states of the NMRDP and the states of the
automaton.
This transformation is of particular interest in
RL with temporally specified rewards expressed in \LTLf/\LDLf, since
it allows to do RL on an equivalent MDP whose (optimal) policies
are also (optimal) policies for the original problem, and viceversa
\cite{BDP-AAAI18}.
This provides the basis for our development here.


In this paper, we set the framework for the problem of restraining
bolt in RL context and provide proofs and practical evidence, through
various examples, that an RL agent can learn policies that optimize
conformance to the \LTLf/\LDLf restraining specifications, without
including in the agent's state space representation the features
needed to evaluate the \LTLf/\LDLf formula.
Our work can also be seen as a contribution to the research providing safety guarantees to  AI techniques based on learning. We take up this point in a brief discussion at the end of the paper.

\section{Preliminaries}

\paragraph{MDP's.}


A Markov Decision Process (MDP) $\M = \langle S,A,Tr,R\rangle$
contains a set $S$ of states, which in this paper we consider factored
into world features, a set $A$ of actions, a transition function
$Tr:S\times A\rightarrow Prob(S)$ that returns for every state $s$ and
action $a$ a distribution over the next state, and a reward function
$R:S\times A \times S \rightarrow\mathbb{R}$ that specifies the reward
(a real value) received by the agent when transitioning from state $s$
to state $s'$ by applying action $a$.
A solution to an MDP is a function, called a {\em policy}, assigning
an action to each state, possibly conditioned on past states and
actions.  The {\em value} of a policy $\rho$ at state $s$, denoted
$v^{\rho}(s)$, is the expected sum of (possibly discounted by a factor $\gamma$, with $0\le\gamma\ 1$) rewards
when starting at state $s$ and selecting actions based on $\rho$.

RL is the task of learning a possibly optimal policy, from an initial 
state $s_0$, on an MDP where only $S$ and $A$ are known, while 
$Tr$ and $R$ are not. 
Typically, the MDP is assumed to start in an initial state $s_0$, so 
policy optimality is evaluated wrt $v^{\rho}(s_0)$.
Every MDP has an {\em optimal} policy $\rho^*$. In discounted
cumulative settings, there exists an optimal policy that is
\emph{Markovian} $\rho: S \rightarrow A$, i.e., $\rho$ depends only
on the current state, and deterministic~\cite{Puterman94}.


\paragraph{LTL$_f$/LDL$_f$.}

The logic \LTLf is the classical linear time logic \LTL \cite{Pnueli77}
interpreted over finite traces, formed by a finite (instead of
infinite, as in \LTL) sequence of propositional interpretations\cite{DegVa13}.
Given a set $\Prop$
of boolean propositions, here called \emph{fluents} \cite{Reiter01}, \LTLf formulas $\varphi$ are defined as follows:
\[\begin{array}{rcl}
\varphi &::=& \phi \mid \lnot \varphi \mid \varphi_1\land \varphi_2 \mid \Next\varphi \mid \varphi_1\Until\varphi_2
\end{array}
\]
where $\phi$ is a propositional formula over $\Prop$, $\Next$ is the
\emph{next} operator and $\Until$ is the \emph{until} operator.
We use the standard abbreviations:
$\varphi_1\lor\varphi_2 \doteq \lnot(\lnot \varphi_1\land \lnot
\varphi_2)$;
\emph{eventually} as $\Diamond\varphi \doteq \true\Until\varphi$;
\emph{always} as $\Box\varphi \doteq\lnot\Diamond\lnot\varphi$; 
week next $\Wnext\varphi \doteq \lnot\Next\lnot\varphi$ (note that on finite
traces $\lnot\Next\varphi \not\equiv \Next\lnot\varphi$); and $Last \doteq \Wnext\false$ denoting the end of the trace. 
\LTLf is as expressive as 
first-order logic (\FO)
over finite traces
and star-free regular expressions (\REGEX).

\LDLf is a proper extension of \LTLf, which is as expressive as
monadic second-order logic (\MSO) over finite traces
\cite{DegVa13}. \LDLf allows for expressing regular expressions over
such sequences, hence mixing procedural and declarative
specifications, as advocated in some work in Reasoning about Action
and Planning \cite{LRLLS97,BaierFBM08}.
Formally, \LDLf formulas $\varphi$ are built as follows:
\[\begin{array}{lcl}
\varphi &::=& \ttrue  \mid \lnot \varphi \mid \varphi_1 \land \varphi_2 \mid \DIAM{\varrho}\varphi \\
\varrho &::=& \phi \mid \varphi? \mid  \varrho_1 + \varrho_2 \mid \varrho_1; \varrho_2 \mid \varrho^*
\end{array}
\]
where: $\ttrue$ stands for logical true; $\phi$ is a propositional
formula over $\Prop$; $\varrho$ denotes path expressions, i.e., \REGEX over
propositional formulas $\phi$ with the addition of the test construct
$\varphi?$ typical of Propositional Dynamic Logic (\PDL).  We use abbreviations
$\BOX{\varrho}\varphi\doteq\lnot\DIAM{\varrho}{\lnot\varphi}$ as in \PDL.
Intuitively, $\DIAM{\varrho}\varphi$ states that, from the current step
in the trace, there exists an execution satisfying the \REGEX $\varrho$ 
such that its last step satisfies $\varphi$, while
$\BOX{\varrho}\varphi$ states that, from the current step, all executions
satisfying the \REGEX $\varrho$ are such that their last step
satisfies $\varphi$.
Tests are used to insert into the execution path checks for
satisfaction of additional \LDLf formulas.

For an \LTLf/\LDLf formula $\varphi$, we can construct a deterministic
finite state automaton (\DFA)~\cite{RaSc59} $\A_\varphi$ that tracks
satisfaction of $\varphi$: $\A_\varphi$ accepts a finite trace $\pi$
{\em iff} $\pi$ satisfies $\varphi$.\footnote{An analogous
  transformation to automata applies to several other formalisms for
  representing temporal specifications over finite traces, including
  \emph{Past \LTL}, \emph{co-safe \LTL}, etc.
  \cite{BBG96,ThiebauxGSPK06,Slaney05,Gretton07,Gretton14,LacerdaPH14,LacerdaPH15}.
  All results presented here apply to those formalisms as well.} 

\newcommand{\Rnm}{\bar{R}}
\newcommand{\rhonm}{\bar{\rho}}

\paragraph{NMRDP's.}

A non-Markovian reward decision process (NMRDP) \cite{BBG96} is a tuple $M=\langle S,A,Tr, \Rnm\rangle$,
where $S,A$ and $Tr$ are as in an MDP, but the reward $\Rnm$ is a real-valued function 
over finite state-action sequences (referred to as \emph{traces}),
i.e., $\Rnm:(S\times A)^* \rightarrow \mathbb{R}$. 
Given a (possibly infinite) trace $\pi=\langle s_0,a_1,\ldots,s_{n-1},a_n\rangle$, the {\em value\/} of $\pi$ is:
$v(\pi) =\sum_{i=1}^{|\pi|} \gamma^{i-1} \Rnm (\langle \pi(1),\pi(2),\ldots,\pi(i)\rangle),$ where
$0<\gamma\leq 1$ is the discount factor and $\pi(i)$ denotes the pair $(s_{i-1},a_i)$. 
In NMRDP's, policies are also non-Markovian $\rhonm: S^*\rightarrow A$.
Since every policy 
induces a distribution over the set of possible infinite traces, we can
define the value of a policy $\rhonm$, given an initial state $s$, as:
$v^{\rhonm}(s) = E_{\pi\sim M,\rhonm,s} v(\pi).$ 
That is, $v^{\rhonm}(s)$ is the expected value of infinite traces,
where 
the distribution over traces is defined by the initial state $s$, the transition function $Tr$, 
and the policy $\rhonm$. 

Specifying a non-Markovian reward function explicitly is cumbersome and unintuitive, 
even if only a finite number of traces are to be rewarded. 
\LTLf/\LDLf provides an intuitive and convenient language rewards
\cite{CamachoCSM17,BDP-AAAI18}. Following, \cite{BDP-AAAI18} we can
specify $\Rnm$ implicitly, using a set of pairs
$\{(\varphi_i,r_i)\}_{i=1}^{m}$, with $\varphi_i$ an \LTLf/\LDLf
formula selecting the traces to reward, and $r_i$ the reward assigned
to those traces, where the atomic propositions, i.e., the fluents, of
$\varphi_i$ correspond to boolean features, or boolean
propositions (e.g., relational value comparison) over the world
features, forming the components of the state vector.
Intuitively, if the current (partial)
trace is $\pi=\langle s_0,a_1,\ldots,s_{n-1},a_n\rangle$, the agent receives at $s_n$ a reward  $r$ if $\varphi_i$ is satisfied by $\pi$. 
Formally, $\Rnm(\pi) = r$ if $\pi\models\varphi$ and $\Rnm(\pi) = 0$, otherwise.




\section{NMRDP with \LTLf /\LDLf rewards}

Before looking at the restraining bolt problem, we review
RL for NMRDP's with \LTLf /\LDLf rewards.


In \cite{BDP-AAAI18} it is shown that for any NMRDP
$M=\langle S,A,Tr,\{(\varphi_i,r_i)\}_{i=1}^{m} \rangle$, there exists
an MDP $M'=\langle S',A,Tr',R'\rangle$ that is \emph{equivalent} to
$M$ in the sense that the states of $M$ can be (injectively) mapped
into those of $M'$ in such a way that corresponding (under the
mapping) states yield same transition probabilities, and corresponding
traces have same rewards \cite{BBG96}.
Denoting with 
$\A_{\varphi_i} = \tup{2^\P, Q_i, q_{i0}, \delta_i,F_i}$ 
(notice that $S\subseteq 2^{\Prop}$ and $\delta_i$ is total) 
the \DFA associated 
with $\varphi_i$, the equivalent MDP $M'$ 
is as follows:
\begin{itemize}\itemsep=0mm
\item $S'=Q_1\times\cdots\times Q_m\times S$;
\item $\Tr' : S'\times A \times S'\rightarrow [0,1]$ is defined as:
\[
\begin{array}{l}
\Tr'(q_1,\ldots,q_m, s, a, q'_1,\ldots,q'_m, s') = {}\\
\quad\left\{
\begin{array}{ll}
Tr(s,a,s') &\mbox{if } \forall i:\delta_i(q_i,s') = q'_i\\
0 & \mbox{otherwise}; 
\end{array}\right.
\end{array}
\] 
\item $R': S'\times A \times S'\rightarrow 
\mathbb{R}$ is defined as:
\[
R'(q_1,\ldots,q_m, s, a, q'_1,\ldots,q'_m, s') = 
\sum_{i: q_i'\in F_i} r_i
\] 
\end{itemize}
Observe that the state space of $M'$ is the product of the state
spaces of $M$ and $\A_{\varphi_i}$, and that the reward $R'$ is
Markovian. In other words, the (stateful) structure of the \LTLf/\LDLf
formulas $\varphi_i$ used in the (non-Markovian) reward of $M$ is
\emph{compiled} into the states of $M'$.
\begin{theorem}[\cite{BDP-AAAI18}]\label{th:main}
The NMRDP $M= \langle S,A,Tr, \{(\varphi_i,r_i)\}_{i=1}^{m} \rangle$ is equivalent to the  MDP $M'= \langle S',A,Tr',R'\rangle$ defined above.
\end{theorem}

Actually this theorem can be refined, into a stronger lemma that we
will use in the following.  A policy $\rho$ for an NMRDP $M$ and a
policy $\rho'$ for an equivalent MDP $M'$ are \emph{equivalent} if
they \emph{guarantee the same rewards}.
Assume $M'$ is constructed as above and let $\rho'$ be a policy for $M'$. 
Consider a trace $\pi=\langle s_0,a_1,s_1,\ldots,s_{n-1},a_n\rangle$ of $M$ 
and assume it leads to state $s_n$. 
Further, let $q_i$ be the state of $\A_{\varphi_i}$ on input $\pi$. 
We define the (non-Markovian) policy $\rhonm$ equivalent to  
$\rho'$ as $\rhonm(\pi)=\rho'(q_1,\ldots,q_m,s_n)$.
Similarly, given a policy $\rho$ for $M$, by just tracking the state of
the \DFAs $\A_{\varphi_i}$, it is immediate to define the equivalent
policy $\rho'$ for $M'$. Hence we have:
\begin{lemma}[\cite{BDP-AAAI18}]\label{th:main-policy}
Given an NMRDP $M$ and an equivalent MDP $M'$, every 
policy  $\rho'$ for $M'$ has an equivalent  policy $\rhonm$ for $M$ and viceversa.\footnote{A variant of this lemma, talking about optimal policy only as originally presented in \cite{BBG96}.}
\end{lemma}

\section{RL for NMRDP with \LTLf /\LDLf rewards}

We are interested in RL in the setting introduced above: learn a
(possibly optimal) policy for an NMRDP
$M=\langle S,A,Tr,\{(\varphi_i,r_i)\}_{i=1}^{m} \rangle$, whose
rewards $r_i$ are offered on traces specified by \LTLf /\LDLf formulas
$\varphi_i$ and where
the
\LTLf /\LDLf reward formulas $\{(\varphi_i,r_i)\}_{i=1}^{m}$ and
the transitions function $Tr$ is hidden to the learning agent.%
\footnote{Observe that $\varphi_i$ is over all world
  features, since we do not distinguish agent features from
  restraining bolt features yet.}
Formally, given $M$, with 
$Tr$ and  $\{(\varphi_i,r_i)\}_{i=1}^{m}$ hidden to the learning agent but sampled during learning, 
and an initial state  $s_0\in S$, the RL problem over $M$
consists in learning an optimal policy $\rhonm$. 
Notice that, since NMRDP rewards are based on traces, instead of
state-action pairs, typical learning algorithms, such as Q-learning
or SARSA \cite{SuttonBarto98}, which are based on MDPs, are not
applicable.
However, by the above results an optimal policy for $M$
can be obtained by learning, instead, an optimal policy for
$M'$. Being $M'$ an MDP, this can be done by typical algorithms such
as Q-learning or SARSA.
Of course, neither $M$ nor $M'$
are (completely) known to the learning agent, and the transformation is 
never done explicitly. Rather, during the learning process, the agent
assumes that the underlying model has the form of $M'$ 
instead of that of $M$.

\begin{theorem}
  RL for \LTLf /\LDLf rewards over an NMRDP
  $M=\langle S, A, Tr, \{(\varphi_i,r_i)\}_{i=1}^{m}\rangle$, with
  $Tr$ and $\{(\varphi_i,r_i)\}_{i=1}^{m}$ hidden to the learning
  agent can be reduced to RL over the MDP
  $M'=\langle S',A,Tr',R'\rangle$ defined above, with $Tr'$ and $R'$
  hidden to the learning agent.
\end{theorem}

Note that $S'$ contains encoding of the stage of satisfaction of the formulas $\varphi_i$. However since the transition function $Tr'$ is hidden, the agent cannot anticipate the effect of an action before the action is executed.

\section{RL with  \LTLf /\LDLf restraining
  specifications}\label{sec:separating}

\begin{figure}
	\centering
	\includegraphics[height=3.1cm]{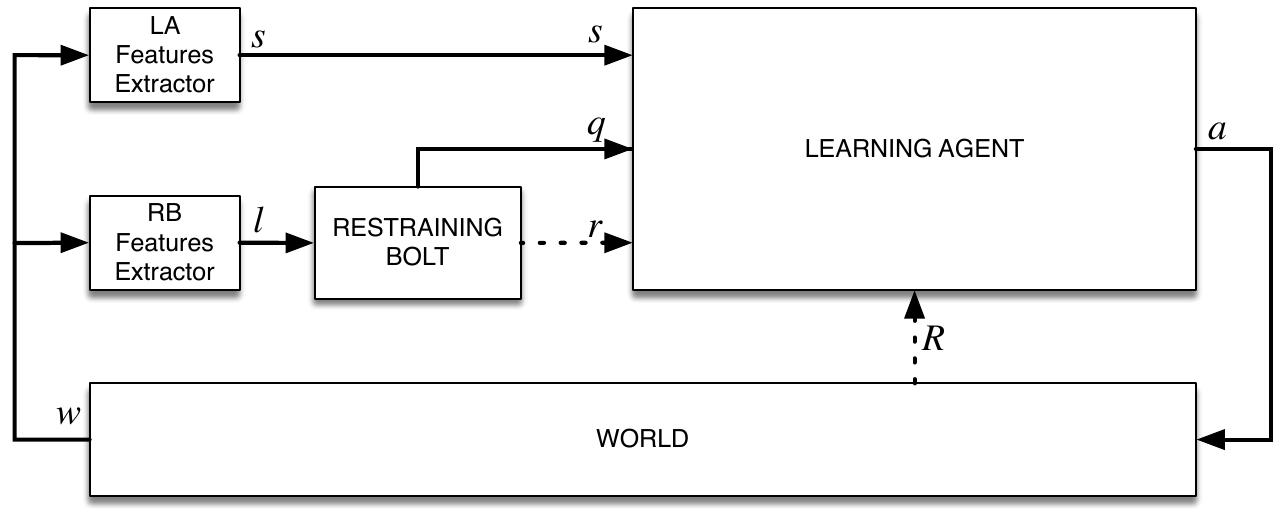} 
	\caption{Learning Agent and Restraining Bolt}
	\label{fig:achitecture2}
\end{figure}

We now focus on the restraining bolt problem, i.e.,  
how to do RL with restraining specifications expressed in \LTLf /\LDLf.

We are given:
\begin{itemize}
	\item A \textbf{learning agent} modeled by the MDP
		  $M_{ag} = \langle S, A, Tr_{ag}, R_{ag} \rangle$, 
		  with transitions $Tr_{ag}$  and rewards $R_{ag}$ hidden, but
		  sampled from the environment.
			
                \item A \textbf{restraining bolt} 
                  $RB =\langle \L,\{(\varphi_i,r_i)\}_{i=1}^{m}
                  \rangle$  where:
		\begin{itemize}
                \item $\L=2^\F$ is the set of possible fluents'
                  configurations (analogously to $S$ denoting the set
                  of configurations of the features available to
                  $M_{ag}$).  Fluents in $\F$ are not among the
                  features that form the states $S$ of the learning
                  agent $M_{ag}$.
			\item $\{(\varphi_i,r_i)\}_{i=1}^{m}$ is a set of \emph{restraining specifications} with
                          \begin{itemize}
                            \item $\varphi_i$, an \LTLf/\LDLf formula over $\F$.  
			 Each $\varphi_i$ selects sequences of fluents' configurations
			  $\ell_1,\cdots,\ell_n$ ($\ell_k\in \L$) whose relationship with 
			  the sequences of states $s_1,\ldots, s_n$ ($s_k \in S$) of
			 $M_{ag}$ is unknown.
			\item $r_i$, the reward associated with $\varphi_i$.
			  A reward $r_i$ is assigned to sequences of
			  configurations $\ell_1,\cdots,\ell_n$ satisfying $\varphi_i$.
                        \end{itemize}
                      \end{itemize}
                    \end{itemize}
                    The agent receives rewards based on $R_{ag}$ and
                    the pairs $(\varphi_i,r_i)$.
                   In fact, often we have to handle tasks of episodic nature. That is, the world can reach a configuration in which no action can change its configuration nor generate new rewards, e.g., a final configuration in a game. In this case we assume that the restraining bolt fluents $\F$ include a special fluent $Done$ that denotes reaching the final configuration. This fluent can be used   in \LTLf/\LDLf formulas to reward the agent only at the end of the episode.
When the episode ends and a new episode is started, a new trace is generated on which \LTLf/\LDLf formulas are evaluated again.

Notice that while the agent can see the features that the reward $R_{ag}$ depends on, 
it cannot see those that affect $r_i$.
Both $S$ and $\L$ are features' configurations, in the sense of
representing world properties. 
However, they capture different facets of the world.
Let $W$ be the set of \emph{real world states}.
A \emph{feature} is a function $f_j : W \rightarrow D_j$ 
that maps world states to another domain $D_j$, 
such as reals, enumerations, booleans, etc.
The \emph{feature vector} of a world state $w_h$ is the vector
$\mathbf{f}(w_h) = \langle f_1(w_h), \ldots, f_d(w_h) \rangle$ of
feature values corresponding to $w_h$.
Given a world state $w_h$, the corresponding 
\emph{configuration $s_h$ of the learning agent $M_{ag}$} 
consists in those components of  $\mathbf{f}(w_h)$ 
that produce the agent's state, while the
corresponding \emph{configuration of fluents} $\ell_h$ is formed by
the components that assign truth values to the fluents.
That is, a subset of the world features describes the agent states $s_h$ and another  
subset (for simplicity, assumed disjoint from the previous one) is used to evaluate 
the fluents in $\ell_h$.
Hence, a sequence $w_1,\ldots,w_n$ of world states defines both a
sequence of learning agent states $s_1,\ldots,s_n$ and a sequence of
fluent configurations $\ell_1,\ldots,\ell_n$.  While $R_{ag}$ depends
on the former, each $\varphi_i$ and $r_i$ depend on the
latter. Consequently, by executing a policy and hence by repeatedly
choosing actions in $A$, the agent visits a sequence of world states,
collecting for each of them, the sum of the rewards $R_{ag}$ and
$r_i$.
The point to resolve is defining on the base of which observations the
agent can choose its next actions. Obviously the agent can in
principle accumulate all its observations $s_1,\ldots,s_n$ and on the
other hand it cannot see the fluents configurations
$\ell_1,\ldots,\ell_n$, however we want to equip the agent to some
means to establish the stage of satisfaction of the formulas
$\varphi_i$. Such a notion, as mentioned above, can be captured by
considering \emph{the} minimal \DFA
$\A_{\varphi_i} =\tup{2^\P, Q_i, q_{i0}, \delta_i,F_i}$ corresponding
to formula $\varphi_i$. Notice that such a \DFA is unique. Hence we
equip the agent with additional observable features
$Q_1\times\ldots\times Q_m$ corresponding to the states Let
$\A_{\varphi_i}$. Such features are going to be provided by the
restraining bolt.\footnote{Notice that coming up with the $Q_1,\ldots,Q_n$ and assigning the rewards to some of them, while can perhaps be possible in very simple cases, without a principled and systematic technique as the one presented here it is virtually impossible.  Indeed, to express directly \LTLf/\LDLf  properties in the MDP one may need exponential additional features, assuming a factorized representation, since the corresponding \DFA may be doubly exponential in the formula.}
Note that this does not give away fluents
configurations $\ell_1,\ldots,\ell_n$ which remain hidden to the
agent, see Figure\ref{fig:achitecture2}

Hence, in general, we consider possibly non-Markovian policies of the form
 \[\rhonm:(Q_1\times\ldots\times Q_m\times S)^*\rightarrow A\]
 and thus define the expected (discounted) cumulative reward of a
 possibly non-Markovian policy $\rhonm$ as the expected reward of
 infinite traces starting in the initial state $s_0$, induced by the
 policy itself (obtained as the expected sum of the collected rewards
 $R_{ag}$ and $r_i$).

\medskip
\noindent
\textbf{Problem definition.} 
(An instance of) the \emph{RL problem with \LTLf/\LDLf restraining specifications} 
is a pair $M_{ag}^{rb}= \langle M_{ag} , RB\rangle$, where:
    $M_{ag} = \langle S, A, Tr_{ag}, R_{ag}
    \rangle$ is a learning agent with $Tr_{ag}$ and $R_{ag}$ hidden, 
    and $RB =\langle \L,\{(\varphi_i,r_i)\}_{i=1}^{m}
    \rangle$ is a restraining bolt formed by a set of \LTLf
    /\LDLf formulas $\varphi_i$ over
    $\L$ with associated rewards
    $r_i$. 
    A solution to the problem is a (possibly non-Markovian) policy 
	$\rhonm:(Q_1\times\ldots\times Q_m\times S)^*\rightarrow A$ that maximizes
	the expected cumulative reward.

\medskip To devise a solution technique, we assume that the agent
actions in $A$ induce a transition distribution over the features and
fluents configuration, i.e.:\footnote{Notice that this assumption
  is quite loose, as we can arbitrarily enlarge $\L$ to define
  $Tr_{ag}^{rb}$. In the construction below only the fluents in $\L$
  that occur in the \LTLf /\LDLf formulas play an active role.}
\[Tr_{ag}^{rb} : S\times \L \times A \rightarrow Prob(S\times \L).\]
Such a transition distribution, together with the initial values of the
fluents $\ell_0$ and of the agent state $s_0$, allow us to describe a
probabilistic transition system accounting for the dynamics of the
fluents and agent states.  Moreover, when $Tr_{ag}^{rb}$ is projected on
$S$ only, i.e., the $\L$ components are marginalized, we get $Tr_{ag}$ of
$M_{ag}$. Obviously, both $Tr_{ag}^{rb}$ and $Tr_{ag}$ are hidden to the learning agent. 
On the other hand, in response to an agent action $a_h$ performed in the current state
$w_h$ (in the state $s_h$ of the agent and the
configuration $\ell_h$ of the fluents), the world changes into $w_{h+1}$ from which $s_{h+1} $ and $\ell_{h+1}$
are obtained.
This is all we need to proceed.

Given $M_{ag}^{rb}= \langle M_{ag} , RB\rangle$ with
$M_{ag} = \langle S, A, Tr_{ag}, R_{ag} \rangle$ and
$RB =\langle \L,\{(\varphi_i,r_i)\}_{i=1}^{m} \rangle$, we define an
NMRDP
$M_{ag}^{n} = \langle S\times \L, A, Tr_{ag}^{rb},
\{(\varphi_i,r_i)\}_{i=1}^{m} \cup \{(\varphi_s, R_{ag}(s,a, s'))\}_{s\in
  S, a\in A, s'\in S} \rangle$, where:
\begin{itemize}
\item states are pairs $(s,\ell)$  formed by an agent configuration $s$ and a fluents configuration $\ell$;
\item $\varphi_i$ are as before;
\item $\varphi_s = \Diamond(s\land a\land \Next(Last \land s'))$;
\item $Tr_{ag}^{rb}$, $r_i$ and  $R_{ag}(s,a, s')$ are hidden and sampled from the environment.
\end{itemize} 

Formulas $\varphi_i$ are as before, in particular they are
continuously evaluated on the (partial) trace produced so far. Though,
they may use the special fluent $Done$ to give the reward
associated to the formula at the end of the episode (modulo reward
shaping).
Formulas $\Diamond(s\land a\land \Next(Last \land s'))$, one per
$(s,a,s')$, which require that both states $s$ and action $a$ are
followed by $s'$, are evaluated at the end of the current (partial)
trace (recall that $Last$ denotes the last element of the
trace, c.f. Preliminaries).  In this case, the reward $R_{ag}(s,a,s')$
from $M_{ag}$ associated with $ (s,a,s')$ is given.\footnote{Notice that we have as many of such formulas as transitions $(s,a,s')$, this causes an exponential blow-up being $S$ factorized in features.  However, we will get rid of them later.}

Observe that policies for $M_{ag}^{n}$ have the form
$(S\times\L)^*\rightarrow A$ which needs to be restricted to have the form required by our problem
$M_{ag}^{rb}$.
A policy $\rhonm:(S\times\L)^*\rightarrow A$ \emph{has the form}
$\rhonm:(Q_1\times\ldots\times Q_m\times S)^*\rightarrow A$ when
$\rhonm(\langle s_1,\ell_1\rangle \cdots \langle s_n,\ell_n\rangle) =
\rhonm(\langle q_{11},\ldots,q_{m1},s_1,\rangle \cdots \langle q_{1n},\ldots,q_{mn},s_n,\rangle)$ with $q_{ij} = \delta_j(\ell_1,\ldots,\ell_i, q_{j0})$.
In other words, a policy  $\rhonm:(S\times\L)^*\rightarrow A$ has the form $\rhonm:(Q_1\times\ldots\times Q_m\times S)^*\rightarrow A$  when the fluents $\L$ are not directly accessible but are used only the progress the \DFAs $\A_{\varphi_i}$ corresponding to formulas $\varphi_i$. 
We can now state the following result.

\begin{lemma}\label{th:goal-nmr}
  RL with \LTLf /\LDLf restraining specifications
  $M_{ag}^{rb}= \langle M_{ag} , RB\rangle$ with
  $M_{ag} = \langle S, A, Tr_{ag}, R_{ag} \rangle$ and
  $RB =\langle \L,\{(\varphi_i,r_i)\}_{i=1}^{m} \rangle$ can be
  reduced to RL over the NMRDP
  $M_{ag}^{n} = \langle S\times \L, A, Tr_{ag}^{rb},
  \{(\varphi_i,r_i)\}_{i=1}^{m}\cup \{(\varphi_s, R_{ag}(s,a, s'))\}_{s\in
    S, a\in A, s'\in S} \rangle$,
    by restricting the policy to learn to have the form $\rhonm:(Q_1\times\ldots\times Q_m\times S)^*\rightarrow A$.
\end{lemma}


\medskip
As a second step, we apply the construction of the previous section and obtain a
new MDP learning agent.  In such construction, because of their
triviality, we do not need to keep track of the state of the automata
associated with each $\varphi_s$, but just offer the reward 
$R_{ag}(s,a,s')$ associated with $(s,a,s')$. Instead, we do need to keep track of 
state of each \DFA $\A_{\varphi_i}=\tup{2^\P, Q_i, q_{i0}, \delta_i,F_i}$ corresponding to $\varphi_i$.
Hence, from $M_{ag}^{n}$, we obtain an MDP $M_{ag}'=\langle S',A,Tr'_{ag},R_{ag}'\rangle$ where:
\begin{itemize}
\item $S'=Q_1\times\cdots\times Q_m\times S\times \L$ is the set of states;
\item $\Tr'_{ag} : S'\times A \times S'\rightarrow [0,1]$ is defined as follows:
\[
\begin{array}{l}
Tr'_{ag}(q_1,\ldots,q_m, s,\ell, a, q'_1,\ldots,q'_m, s',\ell') = {}\\
\quad\left\{
\begin{array}{ll}
Tr_{ag}(s,\ell,a,s',\ell') &\mbox{if } \forall i:\delta_i(q_i,\ell') = q'_i\\
0 & \mbox{otherwise}; 
\end{array}\right.
\end{array}
\] 
\item $R_{ag}': S'\times A \times S' \rightarrow 
\mathbb{R}$ is defined as:
\[\begin{array}{l}
R_{ag}'(q_1,\ldots,q_m, s,\ell, a, q'_1,\ldots,q'_m, s',\ell') = {}\\
\qquad
\sum_{i: q'_i\in F_i} r_i+R_{ag}(s,a,s')
  \end{array}
\] 
\end{itemize}
Observe that, besides the rewards $R_{ag}(s,a,s')$ of the original
learning agent, the environment now offers the rewards $r_i$
associated with the formulas $\varphi_i$, thus guiding the agent
towards the satisfaction of the $\varphi_i$ (by progressing correctly
the corresponding \DFAs $\A_{\varphi_i}$).

By Theorem~\ref{th:main}, it follows that the NMRDP $M_{ag}^{n}$ and
the MDP $M_{ag}'$ are equivalent. Hence, by
Lemma~\ref{th:main-policy}, any policy of $M_{ag}^{n}$ has an
equivalent policy (hence guaranteeing the same reward) in $M_{ag}'$,
and viceversa.  We can thus learn a policy on $M_{ag}'$ instead of
$M_{ag}^{n}$.
We can thus refine Lemma~\ref{th:goal-nmr} into the following.

\begin{lemma}\label{th:goal-mdp-ell}
  RL with \LTLf /\LDLf restraining specifications
 $M_{ag}^{rb}= \langle M_{ag} , RB\rangle$ with
$M_{ag} = \langle S, A, Tr_{ag}, R_{ag} \rangle$ and
$RB =\langle \L,\{(\varphi_i,r_i)\}_{i=1}^{m} \rangle$ can be reduced to RL over the
  MDP $M_{ag}'=\langle S',A,Tr'_{ag},R_{ag}'\rangle$,
  by restricting the policy to learn to have the form $Q_1\times\ldots\times Q_n\times S\rightarrow A$. %
\end{lemma}
This Lemma allows for restricting non-Markovian policies $(Q_1\times\ldots\times Q_n\times S)^* \rightarrow A$ to Markovian policy $Q_1\times\ldots\times Q_n\times S\rightarrow A$ without loss of generality.


\smallskip
As a last step, we
solve the original RL task on $M_{ag}^{rb}$ by performing RL on a new MDP 
$M_{ag}^{q}=\langle Q_1\times\cdots\times Q_m \times S,A,Tr''_{ag},R_{ag}''\rangle$, where:
\begin{itemize}
\item  The transition distribution $Tr''_{ag}$ is the marginalization of $Tr'_{ag}$ wrt $\L$, and is unknown;
\item  The reward $R_{ag}''$ is defined as:\\
\[
R_{ag}''(q_1,\ldots,q_m, s, a, q'_1,\ldots,q'_m, s') {=}\hspace{-1.5ex} \sum_{i: q'_i\in F_i} \hspace{-1.5ex} r_i{+}R_{ag}(s,a,s').
\] 
\item The states $q_i$ of the \DFAs $\A_{\varphi_i}$ are progressed correctly by the environment.
\end{itemize}
Thus, we finally obtain our main result.


\begin{theorem}\label{th:goal-mdp}
  RL with \LTLf /\LDLf restraining specifications
  $M_{ag}^{rb}= \langle M_{ag} , RB\rangle$ with
  $M_{ag} = \langle S, A, Tr_{ag}, R_{ag} \rangle$ and
  $RB =\langle \L,\{(\varphi_i,r_i)\}_{i=1}^{m} \rangle$ can be
  reduced to RL over the MDP
  $M_{ag}^{q}=\langle Q_1\times\cdots\times Q_m \times
  S,A,Tr''_{ag},R_{ag}''\rangle$ and optimal policies
  $\rho_{ag}^{new}$ for $M_{ag}^{rb}$ can be learned by learning
  corresponding optimal policies for $M_{ag}^{q}$.
\end{theorem}



\newcommand{\States}{S}
\newcommand{\Actions}{A}
\newcommand{\TrFun}{Tr}
\newcommand{\Reward}{R_{ag}}
\newcommand{\DiscFact}{\gamma}
\newcommand{\Policy}{\rho}
\newcommand{\ExpRet}{G}
\newcommand{\ValFun}{v}
\newcommand{\qFun}{q}

\newcommand{\bqs}{\vec{q}}
\newcommand{\MDPagent}{M_{ag}}


\noindent\textit{Proof.}
For brevity we use $\bqs$ to denote $q_1, \dots, q_m$.
By Lemma~\ref{th:goal-mdp-ell} we can focus on RL over the
  MDP $M_{ag}'=\langle S',A,Tr'_{ag},R_{ag}'\rangle$ under the restriction that 
  the policy to learn has the form $Q_1\times\ldots\times Q_n\times S\rightarrow A$.

 Notice  that from the definitions of $\Reward'$ and $\Reward''$, we have that for all $\ell, \ell'\in\L$,
 $\Reward'(\bqs,s,\ell, a,\bqs', s', \ell')=\Reward''(\bqs,s,a,\bqs', s') = \sum_{i: q'_i\in F_i}  r_i{+}\Reward(s,a,s').$
	The crux of the proof is to show that for any optimal policy $\Policy$  the values $\ValFun^{\Policy}(\bqs, s,\ell)$ of the state value function for $\M'_{ag}$ do not depend on $\ell$. That is, it is necessary that $\forall \ell_1, \ell_2.v^\Policy(q_1,\ldots,q_m, s, \ell_1) =v^\Policy(q_1,\ldots,q_m, s, \ell_2)$.
	
        To see this, let $\TrFun'_{ag}(s,a,s') = P(s' | s,a)$, then the Bellman equation in our case is:
        $	
	\ValFun^\Policy(\bqs, s, \ell) = 
	\sum_{\bqs', s', \ell'} P(\bqs', s', \ell' | \bqs, s, \ell, a)[\Reward'(\bqs,s,\ell, a,\bqs', s', \ell') + \DiscFact \ValFun^\Policy(\bqs', s', \ell')].
      $
      By using the equality between $\Reward'$ and $\Reward''$ we have:
	$
	\ValFun^\Policy(\bqs, s, \ell)  =
	\sum_{\bqs', s', \ell'} P(\bqs', s', \ell' | \bqs, s, \ell, a)[\Reward''(\bqs,s, a,\bqs', s') + \DiscFact \ValFun^\Policy(\bqs', s', \ell')].
	$
	On the other hand, observe that we can compute $\bqs'$ from $\bqs$ and $\ell'$, that is we do not need $\ell$. Hence:
	$P(\bqs', s', \ell' | \bqs, s, \ell, a) = P(\bqs', s', \ell' | \bqs, s, a).$
        So we can write:
	$\ValFun^\Policy(\bqs, s, \ell) =
	\sum_{\bqs', s', \ell'} P(\bqs', s', \ell' | \bqs, s, a)
	[\Reward''(\bqs,s, a,\bqs', s') + \DiscFact  \ValFun^\Policy(\bqs', s', \ell')].$
	At this point, we see that $\ValFun^\Policy$ does not depend from $\ell$, hence we can safely drop $\ell$ as argument for $\ValFun_\Policy$. Indeed, we get:	$\ValFun^\Policy(\bqs, s) = 
		\sum_{\bqs', s'} [\Reward''(\bqs,s, a,\bqs', s') + \DiscFact  \ValFun^\Policy(\bqs', s')] \sum_{\ell'} P(\bqs', s', \ell' | \bqs, s, a)$
and by  marginalizing the distribution $P(\bqs', s', \ell' | \bqs, s, a)$ over $\ell'$, we get:
             $\ValFun^{\Policy}(\bqs, s)  =
		\sum_{\bqs', s'} P(\bqs', s'| \bqs, s, a)[\Reward''(\bqs,s, a,\bqs', s') + \DiscFact \ValFun^{\Policy}(\bqs', s')].$
        This is Bellman's equation for $M_{ag}^{q}$, hence the 
thesis.
\hfill\qed





This theorem provides us with a technique to learn the optimal policy for RL with \LTLf/\LDLf restraining specification by making minimal intervention to the learning agent: essentially we need to feed it with the rewards $r_i$ at suitable times, and we need to allow the learning agent to keep track of the stage of satisfaction of the restraining bolt formulas by feeding it with new features for $Q_1,\ldots,Q_n$.

\section{Implementation and Examples}


Implementation of agents learning policies with restraining specifications is performed by assuming a learning phase in simulation and an execution phase on the real world.
The learning phase is obtained by combining three software components: 1) a simulator of the dynamic system, 2) a restraining bolt (RB) process, 3) a reinforcement learning (RL) agent. 
All these components are modular (i.e., they can be properly connected each other or replaced by other similar components).
The simulator is responsible for computing the evolution of the dynamic system under study: it receives decisions (actions to be executed) by the RL agent and communicates: i) the current state of the system to both the RL agent and the RB process, and ii) the current reward value to the RL agent.
The RB process receives the current state from the simulator, evaluates the \LTLf/\LDLf formulas denoting the restraining specifications and sends to the RL agent an encoding of the progress of the \DFA representing the formulas and reward values associated to their evolution. Finally, the RL agent receives the simulator state, the RB state, and the rewards and decides the actions to be executed, while computing an optimal policy.
By using such a simulator, the RL agent can learn a policy that maximizes the cumulative discounted reward taking into account both rewards from the environment and rewards from the RB.
In general, when enough training is allowed, the computed policy, when executed on the real world, will satisfy the RB specifications.

As mentioned, the RL agent and the RB process have different sensors to perceive 
different aspects of the state of the world (or of the simulator). So we assume 
that they are implemented with real sensors (when attached to the real world) 
and corresponding virtual sensors (when attached to the simulator). We also 
assume that the simulator is able to model all the relevant evolutions of the 
world that are needed to learn the specific task with restraining 
specifications.


\newcommand\figspace{-.6cm}

\newcommand\h{1.85cm}
\begin{figure}
	\centering
	\includegraphics[height=\h]{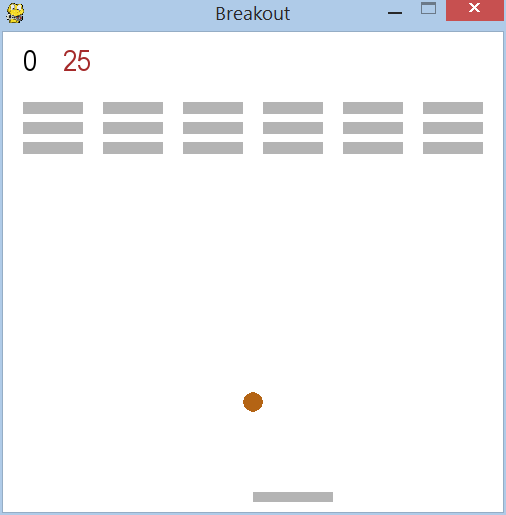}
	\includegraphics[height=\h]{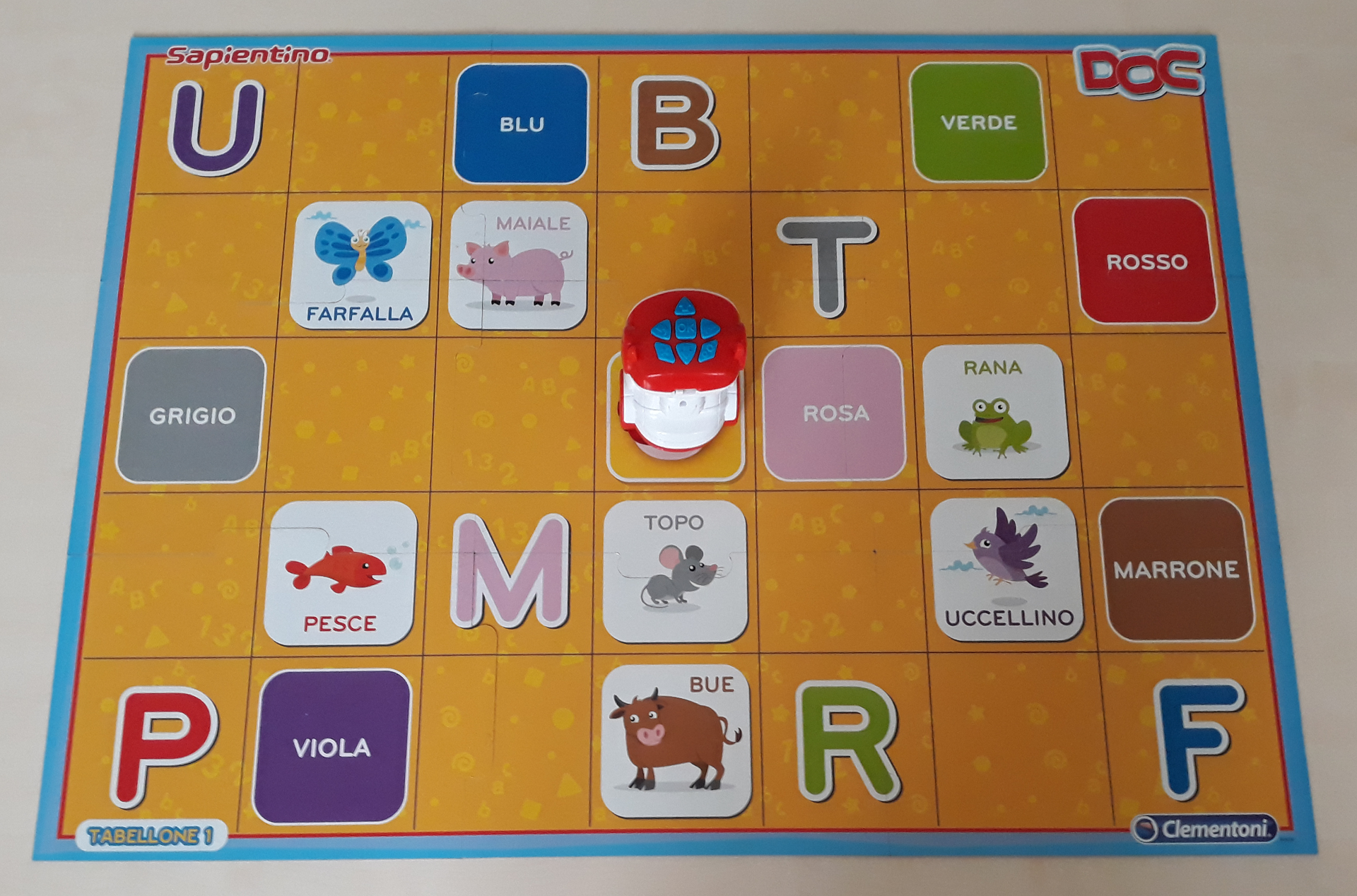}
	\includegraphics[height=\h]{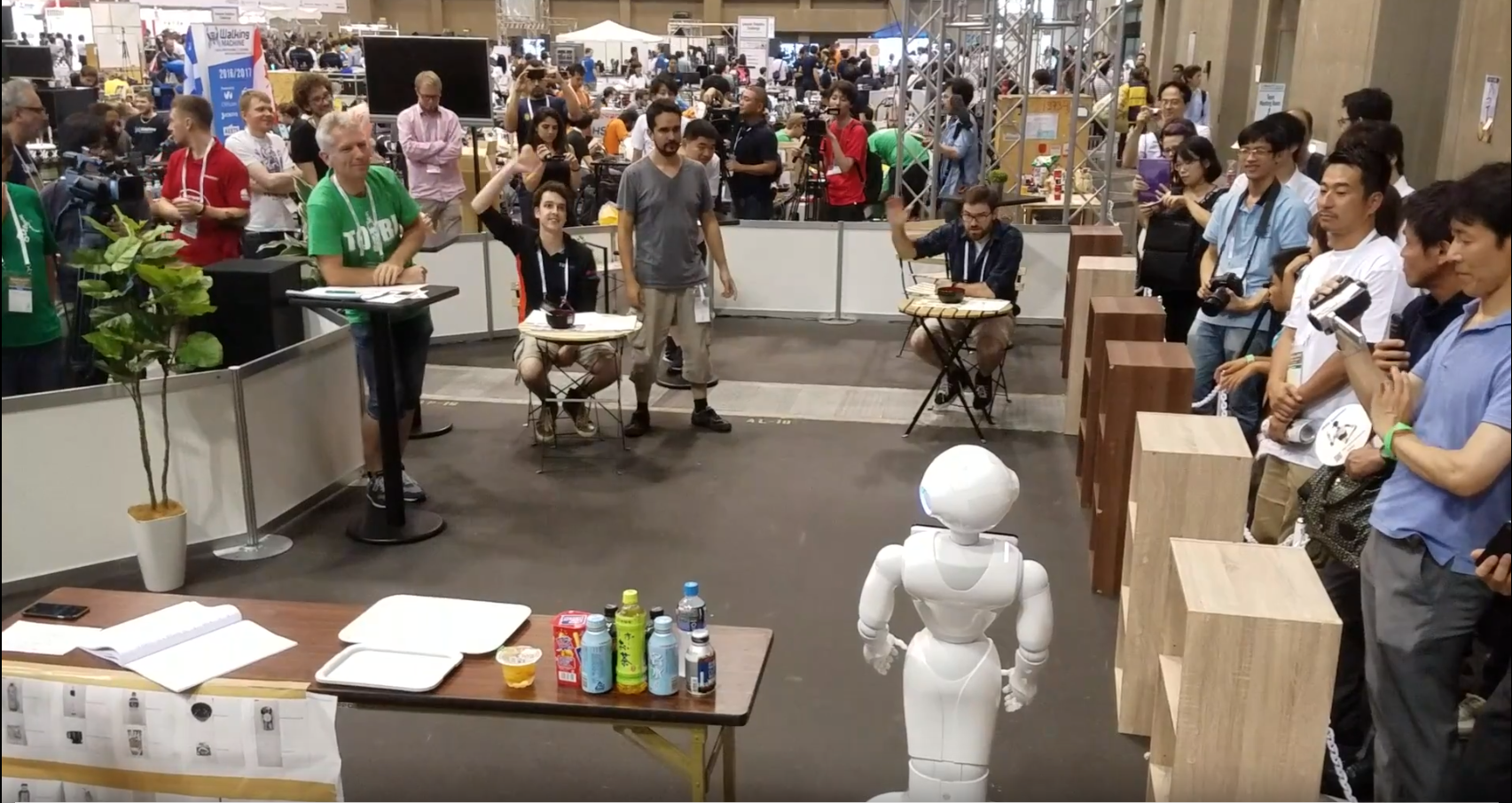} 
	\vspace{.2cm}
	\caption{Experimental scenarios: \Breakout, \Sapientino, \CocktailParty}
	\label{fig:games}
	\vspace{\figspace}
\end{figure}

Next we show the implementation of such components in three 
examples (Figure~\ref{fig:games}).
%
The first one uses a video-game simulator, while the other two 
consider robotic tasks and their corresponding models in a simulator. The core 
software for the RL agent and for the RB process are domain-independent, while 
the (virtual) sensors and the \LTLf/\LDLf specifications are domain-dependent.
Since all examples are of episodic nature, the learning phase is managed by 
an execution system that resets episodes when any of the following conditions 
is verified: 1) a state of the \DFA where the formula is satisfied is reached, 
2) a failure state (i.e., a state from which it is not possible to satisfy any 
formula) of the \DFA is reached, 3) a maximum number of actions have been 
executed (to avoid infinite loops).

To speed up learning, the implementation of the bolt monitors the progress of the \DFA corresponding to the restraining specifications and applies a kind of reward shaping 
by exploiting the \DFA structure\footnote{Reward shaping is not described here 
for lack of space}. Through reward shaping we can anticipate part of the reward 
coming from temporal specifications without waiting for the formulas to 
become true.

Each experiment (i.e., a sequence of episodes to learn a policy) terminates 
after a time limit that is different for each problem (see next sections) 
and chosen to guarantee that a policy consistent with the 
specifications is always found, 
although in general not optimal.
%
%
All the problems described below have been solved with n-step Sarsa algorithm,
configured with $\gamma = 0.999$, $\epsilon = 0.2$, $n=100$. The trend of the solutions
is anyway not sensitive to these parameters\footnote{See more results on the web site.}.

Algorithms have been implemented as single-thread non-optimized Python procedures, in a modular and abstract way to operate on every problem. 
More details about the experimental configurations,
source code of the implementation allowing for reproducing the results contained in this paper, and videos of the found policies are available in {\url{www.diag.uniroma1.it/restraining-bolt}.

\smallskip\noindent
\textbf{Breakout.}
\Breakout has been widely used to demonstrate RL approaches. 
The goal of the agent is to control the paddle in order to drive a ball to hit all the bricks in the screen.
In this example, we have considered two agents with different abilities:
\Move: the agent moves sideways to bounce the ball; \Move+\Fire: the agent can both move and fire straight up to remove bricks.
Agent's state representation uses the following features: $f_x$:
$x$ position of the paddle; $f_{bx}, f_{by}, f_{dx}, f_{dy}$: position
and direction of movement of the ball\footnote{Other state representations are also suitable to learn the task.}.
Reward is given to the agent when a brick is hit.
With this specification a RL algorithm can find a policy to remove all the 
bricks and complete the game for both the agents.

\smallskip\noindent
\emph{Restraining bolt}. We want to provide the agents with the following 
specification:
\emph{the bricks must be removed from left to right}, i.e., all the bricks in column $i$ must be removed before completing any other column $j > i$.
This specification can be expressed with an \LTLf/\LDLf formula and to evaluate such a formula, the  bolt needs a representation  $f_{r_(i,j)}$ of the status of each brick $r_{i,j}$ (present or removed).
The agents, after receiving in input from the bolt an encoding of the status of the \LTLf/\LDLf formula and associated rewards, can use the same RL algorithm to learn a new policy that will complete the task (i.e., remove all the bricks) following the restraining bolt specification (i.e., from left to right).




%

\newcommand\w{.255}
\begin{figure}
\begin{center}
\subfigure{\includegraphics[scale=\w]{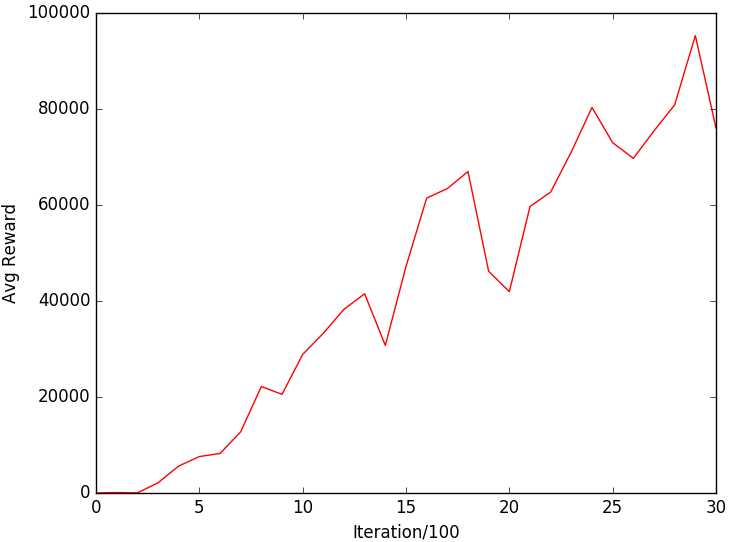}}
\subfigure{\includegraphics[scale=\w]{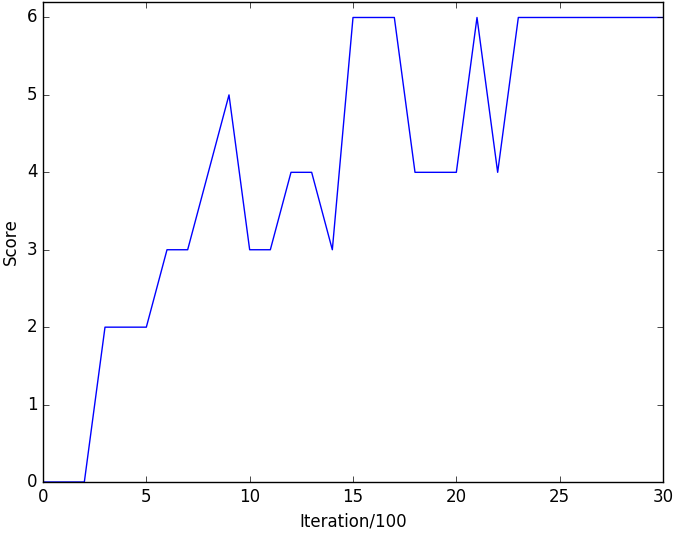}}\\
\subfigure{\includegraphics[scale=\w]{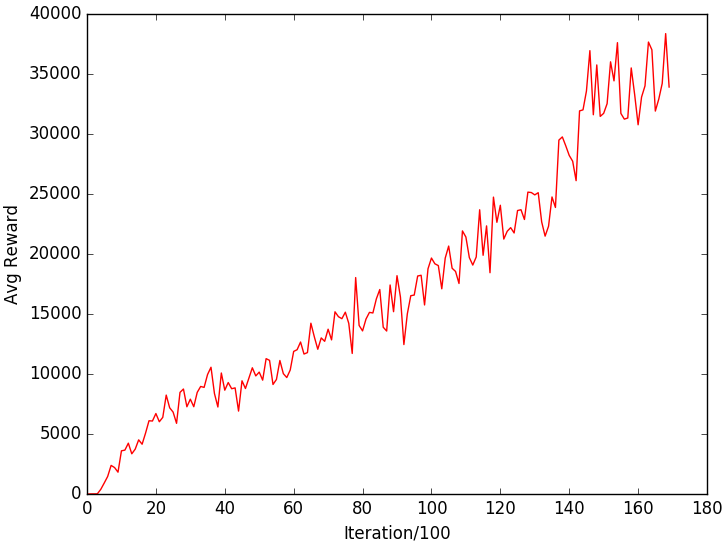}}
\subfigure{\includegraphics[scale=\w]{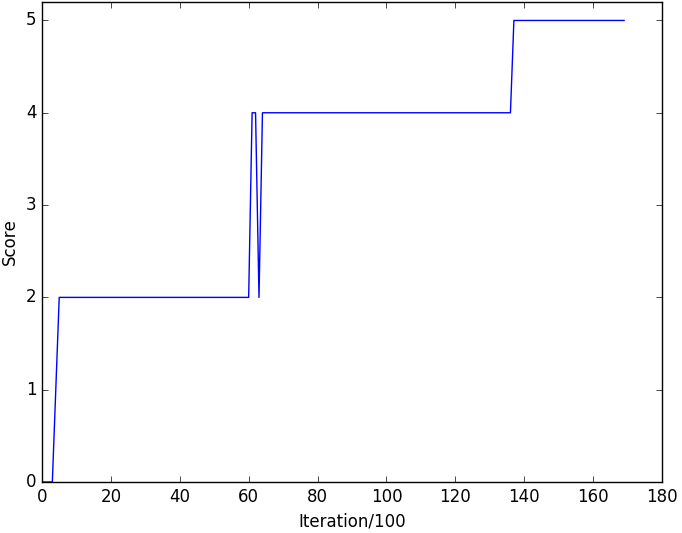}}\\
\subfigure{\includegraphics[scale=\w]{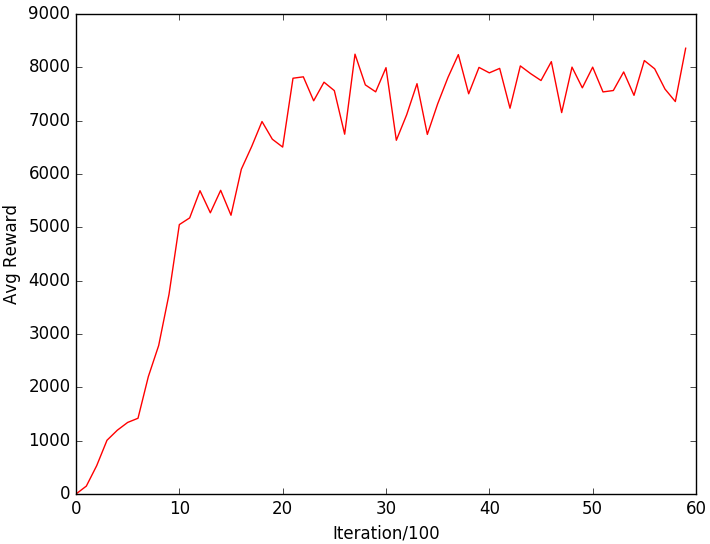}}
\subfigure{\includegraphics[scale=\w]{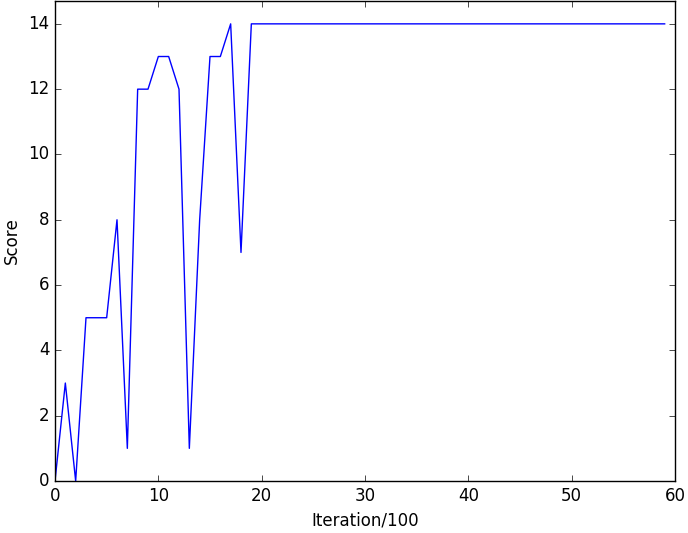}}\\
\subfigure{\includegraphics[scale=\w]{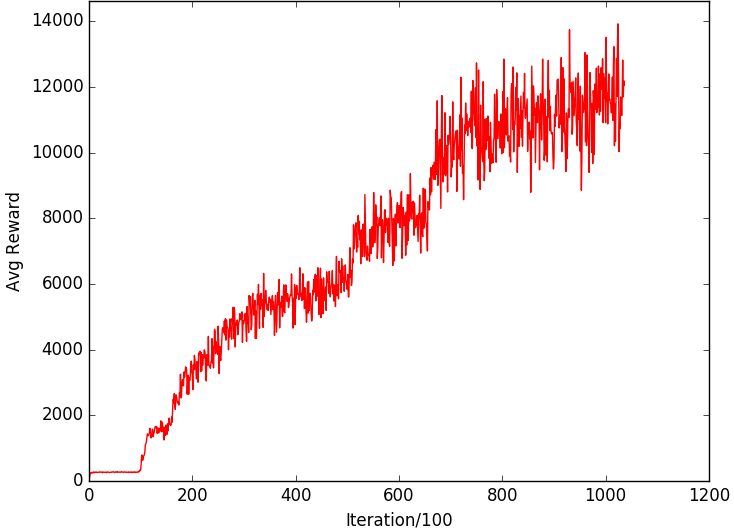}}
\subfigure{\includegraphics[scale=\w]{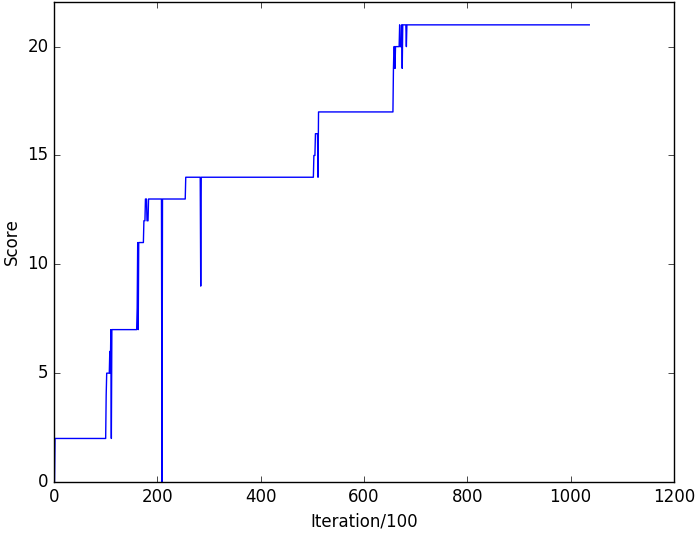}}\\
\subfigure{\includegraphics[scale=\w]{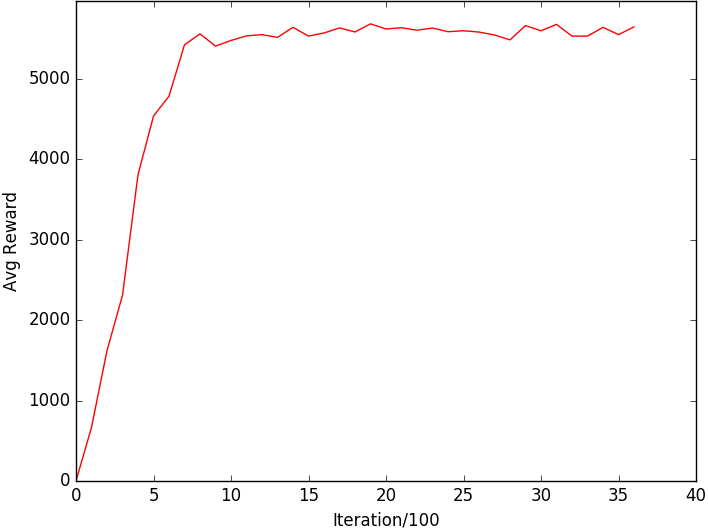}}%
\subfigure{\includegraphics[scale=\w]{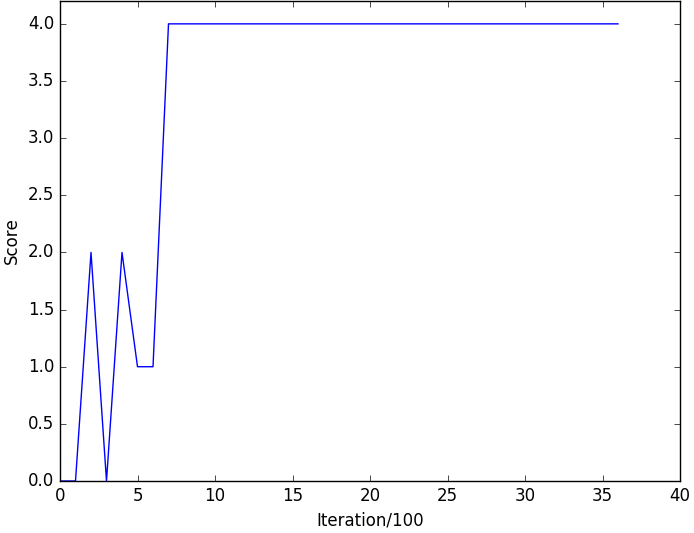}}
\end{center}
\caption{Average reward and scores over number of iterations. Row 1: Breakout \Move + \Fire 4x6 bricks (5 minutes); Row 2: Breakout \Move only 4x5 bricks (1 hour); Row 3:
Sapientino S2 \Omni (3 minutes); Row 4: Sapientino S3 \Differential (1 hour); Row 5: 
Cocktail Party (3 minutes).}
\label{fig:br}
\vspace{-.5cm}
\end{figure}

Notice that the same restraining bolt is applied to the two different agents and 
they will both learn the behavior specified by the \LTLf/\LDLf formula, 
obviously with different policies.
Rows 1 and 2 in Figure \ref{fig:br} show the results of two experiments in the 
Breakout scenario with the following configurations: 
Breakout 4x6 \Move + \Fire (5 minutes), Breakout 4x5 \Move (1 hour).
Left plots show the average reward over the number of iterations, while right plots show the score (i.e., number of columns correctly broken) of the best policy computed so far (i.e., the results obtained in runs without exploration).
The figures show how the agent is able to progressively learn how to progress over the states of the \DFA corresponding to the \LTLf/\LDLf specification.
Similar results are obtained in different configurations (e.g., different sizes of the bricks).
reported in the columns is encoded with the first letter being either M for 
\Move and F for \Fire actions available, and the second letter being either L 
for \Local and G for \Global for the sensor modality.


\smallskip\noindent
\textbf{Sapientino.}
\Sapientino Doc is an educational game for 5-8 y.o. children where a small 
mobile robot has to be programmed to visit specific cells in a 5x7 
grid. Some cells contain concepts that must be matched by the children (e.g., a 
colored animal, a color, the first letter of the animal's name), 
while other cells are empty.
The robot executes sequences of actions given in input by children with a keyboard on the robot's top side. During execution, the robot moves on the grid and executes an action (actually a \emph{bip}) to announce that the current cell has been reached (this is called a \emph{visit} of a cell). A pair of consecutive visits are correct when they refer to cells containing matching concepts. 
As in the real game, we consider a 5x7 grid with 7 triplets of colored cells, each triplet representing three matching concepts.
\emph{State representation} is defined by the following features: $f_x, f_y, f_{\theta}$ reporting the pose of the agent in the grid. 
In this scenario, we consider two different agents: \Omni: omni-directional movements (actions: up, down, left, right), \Differential: differential drive (actions: forward, backward, turn left, turn right).
With this specification, the agent can just learn how to move in the grid, but 
it cannot match related concepts.

\smallskip\noindent
\emph{Restraining bolt}.
Consider now the specifications
S2: \emph{visit at least two cells of the same color for each color, in a given 
order among the colors} (the order of the colors is predefined: first 
$C_1$, then $C_2$, and so on) and
S3: \emph{visit all the triplets of each color, in a given order among the colors}.
%
The following additional features are needed to express and evaluate the 
corresponding formula: $f_{b}$ reporting that a $bip$ action has just been 
executed and $f_{c}$ reporting the color of the current cell.

The restraining specifications for these games can be expressed with \LTLf formulas. A fragment of \LTLf formula for the first game relative to the first color $C_1$ is
\[\begin{array}{l}
\lnot bip \Until (\bigvee_{j=1,2,3} cell_{C_1,j} \land bip) \land {}\\
\bigwedge_{j=1,2,3} \Box(cell_{C_1,j} \land bip \limp \Next\Box(bip \limp \lnot cell_{C_1,j})) \land {}\\
\bigvee_{j=1,2,3} \Box(cell_{C_1,j} \land bip \limp\Next(\lnot bip \Until \bigvee_{k \neq j} cell_{C_1,k} \land bip) \\
\end{array}  
\]
For other colors $C_{i+1}$, we use a similar formula, but requiring that $\bigvee_{j=1,2,3} cell_{C_{i},j} \land bip$ has already been satisfied.

\smallskip


Two agents and two restraining bolts can be combined to form 4 different 
learning situations. We show only two of them.
Rows 3 and 4 of Figure \ref{fig:br} show the agents' learning ability (score = 
14 for S2 specs, score = 21 for the S3 
specs).
Similar results are obtained in different configurations.

\smallskip\noindent
\textbf{Cocktail party.}
For a service robot involved in a cocktail party 
we consider a representation of the state in terms of robot's 
pose and objects' (drinks and snacks) and people's location. 
The agent can move in the environment, grasp and deliver
items to people, and
get a reward when a delivery task is completed. 
The robot has no sophisticated people 
perception capabilities, and no memory
is available in 
the underlying MDP modeling the domain, so the robot cannot
get 
information about individual people or remember who
received what. 
The robot in this scenario will just learn how to bring one
item to any person 
(choosing the shortest path).

\smallskip\noindent
\emph{Restraining bolt}.
Consider the following specification: \emph{serve exactly one drink and 
one snack to every person, but do not serve alcoholic drinks to minors}.
As in the previous examples,
the restraining bolt works on separate features, namely identity, age and received items\footnote{In practice, services like Microsoft Cognitive Services Face API can be integrated into the bolt to provide this information.}
and uses an \LTLf/\LDLf formula to model this specification.
operating scenario. We assume the map of the environment to be known, people 
sitting at tables in predefined known positions and locations of snack and drink 
items also known. From these information we can instantiate a simulator for the 
robot to navigate in this environment and reach the different 
locations\footnote{Specifically, we used Stage simulator in ROS with standard 
navigation stack.}.
%

For learning this task, we considered a problem with two people and two different kinds of drinks and snacks (4 tasks to be executed in total)
and we implemented an abstract simulator reproducing the scenario of RoboCup@Home competition.
The results of learning the restrained task in the simulator are depicted in Row 5 of Figure \ref{fig:br} (score = 4 means that the 2 persons have received one drink and one snack each). As shown, after about 1 minute of simulation\footnote{This time can be drastically reduced using optimized code.}, the RL agent converged to a policy satisfying the RB specifications.

\medskip\noindent
\textbf{Minecraft.}
As an example of our approach's modularity, 
we used the same agent of 
\Sapientino in a \Minecraft scenario. Here the agent has to accomplish 10 
tasks (described with non-Markovian rewards via an \LTLf/\LDLf 
formula).
The two agents share the same state representation $S$ but
differ in the action set 
$A$, the fluent configurations $\L$, and the component progressing the \DFAs.
Results (not shown here) confirm that a \emph{general-purpose} agent can learn 
several tasks by only receiving information from its restraining bolt.

\section{Conclusions}

We have shown how to perform RL with \LTLf/\LDLf
restraining specifications by resorting to typical RL techniques based
on MDPs. Notably, we have shown that the features
needed to evaluate \LTLf/\LDLf formulas can be kept separated from those 
directly accessible to the learning agent.

Our work can be ascribed to that part of research generated by the urgency of 
providing safety guarantees to AI techniques based on learning
\cite{AmodeiOSCSM16,Hadfield-Menell16a,OrseauA16}.
In particular, it shares similarities with recent work on constraining the RL 
agent to satisfy certain safety conditions 
\cite{WenET15,AchiamHTA17,AlshiekhBEKNT18}.
There are however important differences.
First, in enforcing the restraining bolt we consider the learning agent 
essentially as a black box. That is, the restraining bolt does not need to know 
the internals $S$ of the learning agent, and specifies the desired constraints 
using only its world features $\L$. On the other hand, we do not guarantee the 
satisfaction of the restraining bolt constraints during training, as in 
\cite{AchiamHTA17}. In fact, differently from \cite{WenET15,AlshiekhBEKNT18}, 
we do not guarantee the hard satisfaction of constraints even after 
training. After all ``You can’t teach pigs to fly”! and we may very well ask to 
do so in our restraining bolts, being these completely unrestricted in the 
selection of world features and in the kind of formulas they specify over 
such features.
If we want to check formally that the optimal policy satisfies the
restraining bolt specification, we first need to model how actions
affect the restraining bolt's features $\L$, i.e., we need to link
the learning agent's features $S$ to $\L$, and then
we can use, e.g., model checking.
Notably, for doing RL we do not need to specify 
such a link, but we can simply allow the (possibly simulated) world to act 
as the link, in line with what advocated, e.g., in \cite{Brooks91}, and very 
differently from what typically considered in knowledge representation 
\cite{Reiter01}.

Apart from restraining bolts, the interest in having separate representations is 
manifold.
The learning agent feature space can be designed 
separately from the features needed to express the goal, thus 
promoting \emph{separation of concerns} which, in turn, 
facilitates the design, providing for
\emph{modularity} and \emph{reuse} of representations (the same agent can learn from different bolts and the same bolt can be applied to different agents). 
%
Also, a reduced agent's feature space allows for realizing
  \emph{simpler agents} (think, e.g., of a mobile robot platform,
  where one can avoid specific sensors and perception routines), while
  preserving the possibility of acting according to complex declarative specifications
  which cannot be represented in the agent's feature space.
%
We plan to investigate this separation further in the future.
\vspace{-1em}
\paragraph{Acknowledgements}
Work partially funded by Sapienza University of Rome, under projects  
``Immersive Cognitive Environments'' and ``Data-awaRe Automatic 
Process Execution'', and by the European Union's Horizon 2020 
research and innovation program under grant agreement N.~825619.


\bibliographystyle{aaai}
\small
\bibliography{biblio}

\end{document}